%% file: example_paper.tex
\documentclass[10pt,twocolumn,letterpaper]{article}

\usepackage{cvpr}            % To produce the CAMERA-READY version
\input{preamble}
\definecolor{cvprblue}{rgb}{0.21,0.49,0.74}
\usepackage[pagebackref,breaklinks,colorlinks,allcolors=cvprblue]{hyperref}
\usepackage{tocloft}
\usepackage{float}
\usepackage{color}
\def\paperID{16} % *** Enter the Paper ID here
\def\confName{CVPR}
\def\confYear{2025}
\newcommand{\NAME}{SynBY}

\makeatletter
\apptocmd\@maketitle{{\teaser{}}}{}{}
\makeatother
\newcommand{\teaser}{%
\vspace{-10pt}
\centering
\begin{tabular}{cccccccccccc}
     & MoBY & \NAME{} &  & MoBY & \NAME{} &  & MoBY & \NAME{} \\
    \includegraphics[width=0.085\textwidth]{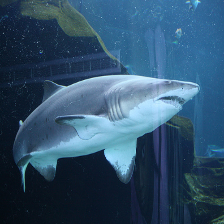} &
    \includegraphics[width=0.085\textwidth]{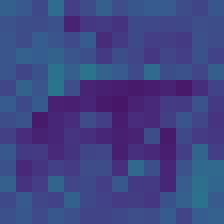} &
    \includegraphics[width=0.085\textwidth]{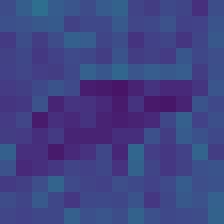} &
    
    \includegraphics[width=0.085\textwidth]{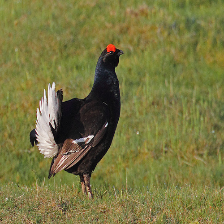} &
    \includegraphics[width=0.085\textwidth]{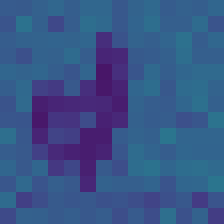} &
    \includegraphics[width=0.085\textwidth]{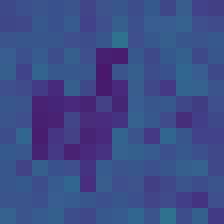} &
    
    \includegraphics[width=0.085\textwidth]{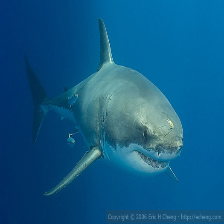} &
    \includegraphics[width=0.085\textwidth]{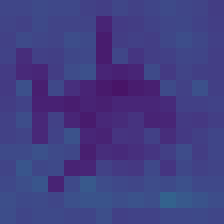} &
    \includegraphics[width=0.085\textwidth]{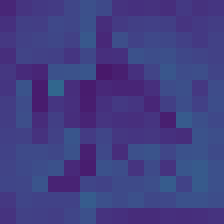}
    
    \\

    \includegraphics[width=0.085\textwidth]{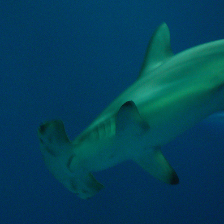} &
    \includegraphics[width=0.085\textwidth]{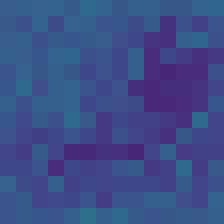} &
    \includegraphics[width=0.085\textwidth]{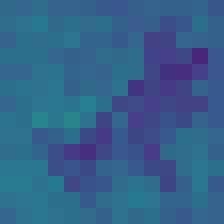} &
    
    \includegraphics[width=0.085\textwidth]{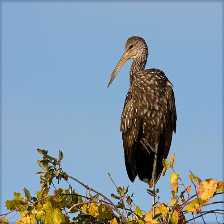} &
    \includegraphics[width=0.085\textwidth]{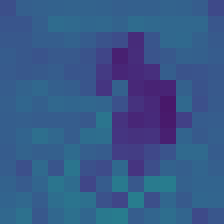} &
    \includegraphics[width=0.085\textwidth]{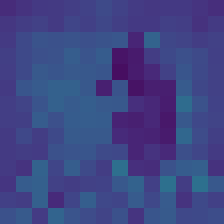} &
    
    \includegraphics[width=0.085\textwidth]{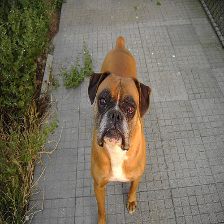} &
    \includegraphics[width=0.085\textwidth]{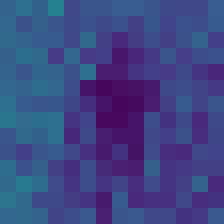} &
    \includegraphics[width=0.085\textwidth]{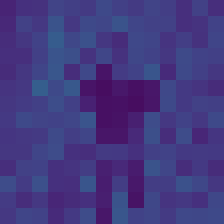}
\end{tabular}\\
\vspace{-6pt}
\captionof{figure}{Self-attention patterns of the average attention head of DeiT-S from the last transformer layer for MoBY and our approach.}
\label{fig:teaser}
\par\vspace{10pt}
}

\title{
Unsupervised Training of Vision Transformers with Synthetic Negatives
}

\author{Nikolaos Giakoumoglou \quad Andreas Floros \quad Kleanthis Marios Papadopoulos \quad Tania Stathaki\\
Imperial College London\\
{\tt\small \{nikos, andreas.floros18, kleanthis-marios.papadopoulos18, t.stathaki\}@imperial.ac.uk}\\
}

\begin{document}
\maketitle

%%%%%%%%%%%%%%%%%%%%%%%%%%%%%%%%%%%%%%%%%%%%%%%%%%%%%%%%%%%%%%%%%%%%%%%%%%%%%%%
%%%%%%%%%%%%%%%%%%%%%%%%%%%%%%%%%%%%%%%%%%%%%%%%%%%%%%%%%%%%%%%%%%%%%%%%%%%%%%%
% 0 ABSTRACT
%%%%%%%%%%%%%%%%%%%%%%%%%%%%%%%%%%%%%%%%%%%%%%%%%%%%%%%%%%%%%%%%%%%%%%%%%%%%%%%
%%%%%%%%%%%%%%%%%%%%%%%%%%%%%%%%%%%%%%%%%%%%%%%%%%%%%%%%%%%%%%%%%%%%%%%%%%%%%%%

\begin{abstract}

This paper does not introduce a novel method per se. Instead, we address the neglected potential of hard negative samples in self-supervised learning. Previous works explored synthetic hard negatives but rarely in the context of vision transformers. We build on this observation and integrate synthetic hard negatives to improve vision transformer representation learning. This simple yet effective technique notably improves the discriminative power of learned representations. Our experiments show performance improvements for both DeiT-S and Swin-T architectures.\footnote{\url{https://github.com/giakoumoglou/synco-v2}}

\end{abstract}

%%%%%%%%%%%%%%%%%%%%%%%%%%%%%%%%%%%%%%%%%%%%%%%%%%%%%%%%%%%%%%%%%%%%%%%%%%%%%%%
%%%%%%%%%%%%%%%%%%%%%%%%%%%%%%%%%%%%%%%%%%%%%%%%%%%%%%%%%%%%%%%%%%%%%%%%%%%%%%%
% 1 INTRODUCTION
%%%%%%%%%%%%%%%%%%%%%%%%%%%%%%%%%%%%%%%%%%%%%%%%%%%%%%%%%%%%%%%%%%%%%%%%%%%%%%%
%%%%%%%%%%%%%%%%%%%%%%%%%%%%%%%%%%%%%%%%%%%%%%%%%%%%%%%%%%%%%%%%%%%%%%%%%%%%%%%

\section{Introduction}\label{sec:introduction}

Computer vision has recently witnessed two major advances. Self-supervised contrastive learning \cite{chen2020simclr, he2020moco} has fundamentally transformed how machines learn from visual data without labels. Concurrently, vision transformer architectures \cite{vaswani2017attention, dosovitskiy2021vit} have reshaped the field by applying attention mechanisms to image understanding tasks. Self-supervised methods have proven remarkably effective for building robust visual representations \cite{liu2021self}, often referred to as \textit{"the dark matter of intelligence"} that underpins broader machine comprehension. As Yann LeCun aptly noted, \textit{"if intelligence is a cake, the bulk of the cake is unsupervised learning"}, and the emergence of transformer models facilitates this by providing architectures capable of capturing complex relationships within visual data \cite{dosovitskiy2021vit}.

Despite their effectiveness, contrastive learning approaches face a persistent challenge regarding the quality of negative examples \cite{kalantidis2020mochi}. Standard techniques rely on randomly sampling negatives from a batch \cite{chen2020simclr, chen2021mocov3} or memory bank \cite{he2020moco, wu2018instdis}, but these negatives are often too easy to distinguish, limiting the discriminative power of learned representations \cite{kalantidis2020mochi, giakoumoglou2024synco}.

In this work, we attempt to overcome this limitation by integrating synthetic hard negatives into self-supervised vision transformer training. Building upon existing momentum-based frameworks \cite{he2020moco,grill2020byol,xie2021moby}, we generate challenging negative examples that force the model to learn more discriminative features \cite{giakoumoglou2024synco, kalantidis2020mochi}. Inspired by recent advances in contrastive learning \cite{giakoumoglou2024synco}, our approach synthesizes hard negatives \textit{"on-the-fly"} in the feature space, creating examples that improve representation quality while maintaining stability. The key insight of our approach is that synthetic negatives provide a controlled way to increase the difficulty of the learning task \cite{giakoumoglou2024synco}, pushing the model to develop more robust representations. 

%%%%%%%%%%%%%%%%%%%%%%%%%%%%%%%%%%%%%%%%%% FIGURE
\begin{figure*}[!t]
    \centering
    \includegraphics[width=0.95\linewidth]{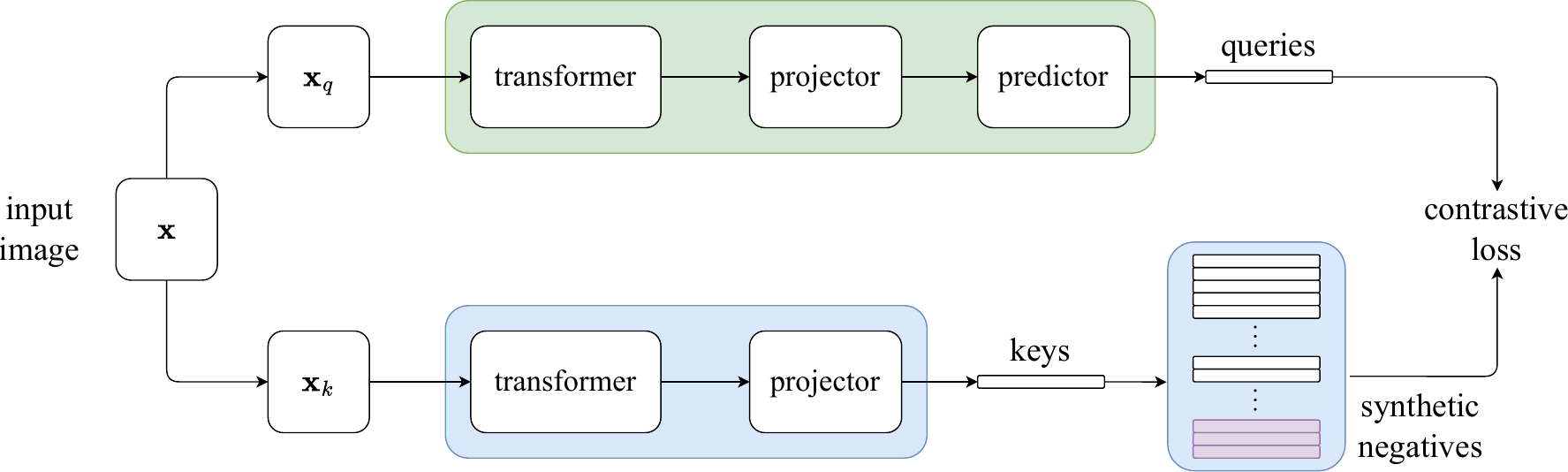}
    \caption{\NAME{} framework overview. Our approach incorporates synthetic hard negatives into the MoBY framework.}
\label{fig:framework}
\end{figure*}
%%%%%%%%%%%%%%%%%%%%%%%%%%%%%%%%%%%%%%%%%%

Our main \textbf{contributions} include exploring the previously uninvestigated application of synthetic negatives in vision transformers. Specifically:

\begin{itemize}
    \item We demonstrate the potential of synthetic hard negatives in contrastive learning by integrating our approach with the MoBY framework and experimenting on the DeiT-S and Swin-T architectures.
    \item We ablate and benchmark on ImageNet, where we find that most configuration settings of our proposed method provide sufficient contrast for the models to learn highly discriminative features.
\end{itemize}

%%%%%%%%%%%%%%%%%%%%%%%%%%%%%%%%%%%%%%%%%%%%%%%%%%%%%%%%%%%%%%%%%%%%%%%%%%%%%%%
%%%%%%%%%%%%%%%%%%%%%%%%%%%%%%%%%%%%%%%%%%%%%%%%%%%%%%%%%%%%%%%%%%%%%%%%%%%%%%%
% 2 RELATED WORK
%%%%%%%%%%%%%%%%%%%%%%%%%%%%%%%%%%%%%%%%%%%%%%%%%%%%%%%%%%%%%%%%%%%%%%%%%%%%%%%
%%%%%%%%%%%%%%%%%%%%%%%%%%%%%%%%%%%%%%%%%%%%%%%%%%%%%%%%%%%%%%%%%%%%%%%%%%%%%%%

\section{Related Work} \label{sec:relatedwork}

\paragraph{Self-supervised visual representation learning.} Unsupervised learning has emerged as a powerful approach to learn visual representations without manual annotations. Within this paradigm, contrastive learning has shown particular promise and has been widely adopted in various forms \cite{chen2020simclr,he2020moco,chen2021mocov3,tian2020infomin}. {SimCLR} \cite{chen2020simclr} demonstrated the effectiveness of a simple framework using data augmentation, large batch sizes, and nonlinear projection heads. {MoCo} \cite{he2020moco} introduced a momentum encoder and queue-based mechanism, enabling contrastive learning with smaller batch sizes.

\paragraph{Hard negatives in contrastive learning.} The quality of negative samples in contrastive learning has been a focus of extensive research \cite{kalantidis2020mochi, robinson2021contrastive, chuang2020debiased, wu2018instdis, arora2019theoretical, giakoumoglou2024synco}. These studies aim to select informative negative samples and address false negatives in instance discrimination tasks. Recent work \cite{kalantidis2020mochi} explored mixing of hard negatives to create challenging contrasts, showing that harder examples lead to improved representations. Subsequent works developed this direction, with newer approaches \cite{giakoumoglou2024synco} proposing systematic methods for generating synthetic hard negatives in the feature space.

\paragraph{Self-supervised transformers for vision.} Self-supervised learning for vision transformers has rapidly evolved \cite{he2021mae, bao2022beit}. Self-distillation methods operate without labels \cite{caron2021dino}, while masked modeling draws inspiration from language processing \cite{he2021mae, bao2022beit}. MoCo-v3 \cite{chen2021mocov3} adapted momentum-based frameworks for transformers, addressing instability through fixed patch projection and batch normalization. Other contrastive methods like MoBY \cite{xie2021moby} implemented asymmetric drop path rates and fewer stability "\textit{tricks}".

%%%%%%%%%%%%%%%%%%%%%%%%%%%%%%%%%%%%%%%%%%%%%%%%%%%%%%%%%%%%%%%%%%%%%%%%%%%%%%%
%%%%%%%%%%%%%%%%%%%%%%%%%%%%%%%%%%%%%%%%%%%%%%%%%%%%%%%%%%%%%%%%%%%%%%%%%%%%%%%
% 3 BACKGROUND
%%%%%%%%%%%%%%%%%%%%%%%%%%%%%%%%%%%%%%%%%%%%%%%%%%%%%%%%%%%%%%%%%%%%%%%%%%%%%%%
%%%%%%%%%%%%%%%%%%%%%%%%%%%%%%%%%%%%%%%%%%%%%%%%%%%%%%%%%%%%%%%%%%%%%%%%%%%%%%%

\section{Background}

In this section, we introduce contrastive learning basics (\Cref{sec:cl}) and our framework for generating synthetic hard negatives (\Cref{sec:synco}), illustrated in \Cref{fig:framework}.

\subsection{Contrastive Learning}
\label{sec:cl}

Contrastive learning aims to learn representations by comparing similar and dissimilar samples. Given an image, $\mathbf{x}$, and two distributions of image augmentations, $\mathcal{T}$ and $\mathcal{T}'$, two augmented views of the same image are created via $\mathbf{x}_q = t_q(\mathbf{x})$ and $\mathbf{x}_k = t_k(\mathbf{x})$, with $t_q \sim \mathcal{T}$ and $t_k \sim \mathcal{T}'$. The views are encoded by \textit{online} and \textit{target} encoders, $f_q$ and $f_k$, respectively, producing vectors \(\mathbf{q}=f_q(\mathbf{x}_q)\) and \(\mathbf{k}=f_k(\mathbf{x}_k)\). The learning objective is to minimize the InfoNCE loss \cite{oord2019cpc}:

\begin{equation}\label{eq:loss_contrastive}
   \mathcal{L}(\mathbf{q},\mathbf{k},\mathcal{Q}) = -\log \frac{\exp(\mathbf{q}^\top \cdot \mathbf{k} / \tau )}{\exp(\mathbf{q}^\top \cdot \mathbf{k} / \tau ) + \sum\limits_{\mathbf{n} \in \mathcal{Q}} \exp(\mathbf{q}^\top \cdot \mathbf{n} / \tau)}.
\end{equation}

\noindent Here, $\mathcal{Q} = \{\mathbf{n}_1, \mathbf{n}_2, \ldots, \mathbf{n}_K\}$ is a set of $K$ negative samples and $\tau$ is a temperature parameter. Negative samples are mined either from the batch \cite{chen2020simclr, chen2021mocov3} or from a memory bank \cite{he2020moco, misra2019pirl, tian2020infomin}. The online encoder is updated via gradient descent whereas the target encoder may be updated via momentum, $\theta_k \leftarrow m \cdot \theta_k + (1-m) \cdot \theta_q$, or through weight sharing in siamese networks (i.e., where $f_k \equiv f_q$).

\subsection{Synthetic Hard Negatives} 
\label{sec:synco}

Synthetic negatives provide challenging examples that help models learn more discriminative features. Let $\mathcal{Q}^N = \text{TopK}(\{\text{sim}(\mathbf{q}, \mathbf{n}) \mid \mathbf{n} \in \mathcal{Q}\}, N)$ be the subset containing the $N < K$ hardest negatives, where $\text{sim}(\cdot, \cdot)$ is the cosine similarity. The synthetic hard negatives can be abstractly represented through a synthesis function, $\mathcal{F}$, as follows:

\begin{equation}\label{eq:synco}
   \mathbf{s}=\frac{\mathcal{F}(\mathbf{q},  \mathcal{Q}^N; \xi)}{\lVert\mathcal{F}(\mathbf{q},  \mathcal{Q}^N; \xi)\lVert_2},
\end{equation}

\noindent where $\lVert \cdot \rVert_2$ denotes the $\ell_2$ norm, and $\xi$ represents the parameters that control the synthesis process, described in \cite{giakoumoglou2024synco}. The set of $L$ synthetic negatives, $\mathcal{Q}_s=\{\mathbf{s}_1, \mathbf{s}_2, \ldots, \mathbf{s}_L\}$, is then combined with the existing queue of real negatives, $\mathcal{Q}$, effectively expanding the diversity of negative examples and exposing the model to more challenging contrasts.

%%%%%%%%%%%%%%%%%%%%%%%%%%%%%%%%%%%%%%%%%% FIGURE 1
\begin{figure*}[!t]
    \centering
    \begin{subfigure}[t]{0.483\textwidth}
        \centering
        \includegraphics[width=\textwidth]{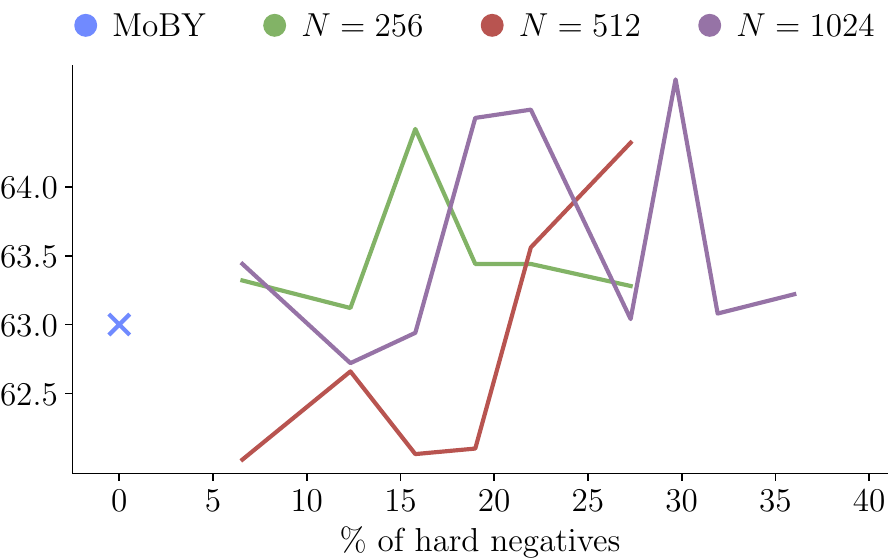}
        \caption{DeiT-S}
         \label{fig:effect_synthetic_negatives_deit}
    \end{subfigure}
    \hfill
    \begin{subfigure}[t]{0.483\textwidth}
        \centering
        \includegraphics[width=\textwidth]{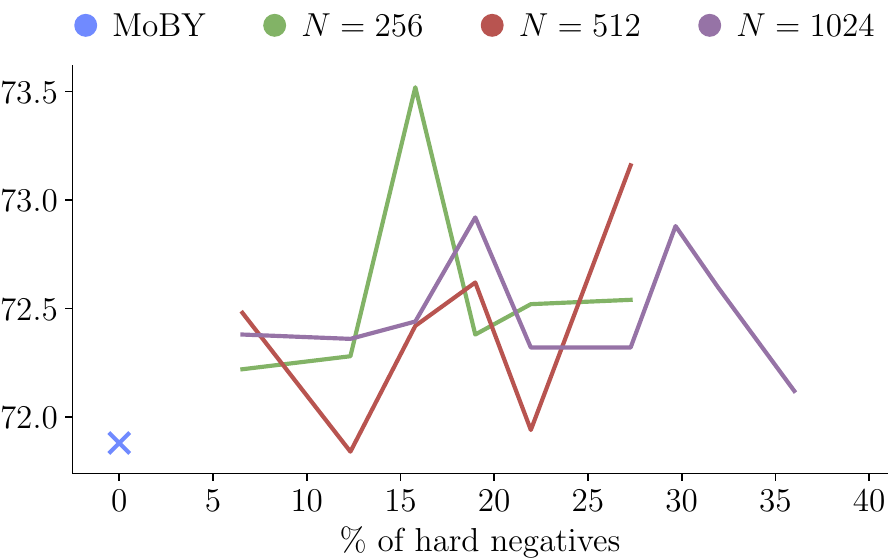}
        \caption{Swin-T}
          \label{fig:effect_synthetic_negatives_swin}
    \end{subfigure}
    \caption{Ablation study of different hardness selection values and synthetic negative percentages.}
    \label{fig:ablation_hardness}
\end{figure*}

%%%%%%%%%%%%%%%%%%%%%%%%%%%%%%%%%%%%%%%%%%%%%%%%%%%%%%%%%%%%%%%%%%%%%%%%%%%%%%%
%%%%%%%%%%%%%%%%%%%%%%%%%%%%%%%%%%%%%%%%%%%%%%%%%%%%%%%%%%%%%%%%%%%%%%%%%%%%%%%
% 4 EXPERIMENTS
%%%%%%%%%%%%%%%%%%%%%%%%%%%%%%%%%%%%%%%%%%%%%%%%%%%%%%%%%%%%%%%%%%%%%%%%%%%%%%%
%%%%%%%%%%%%%%%%%%%%%%%%%%%%%%%%%%%%%%%%%%%%%%%%%%%%%%%%%%%%%%%%%%%%%%%%%%%%%%%

\section{Experiments} \label{sec:experiments}

We develop our approach in PyTorch, building upon the implementation of MoBY \cite{xie2021moby} and SynCo \cite{giakoumoglou2024synco}. We refer to our method as \NAME{}.

\subsection{Setup}

We pretrain \NAME{} on ImageNet ILSVRC-2012 \cite{deng2009imagenet} and its smaller ImageNet-100 subset \cite{khosla2021supcon} using a DeiT-Small \cite{dosovitskiy2021vit,touvron2021deit} or Swin-Tiny \cite{liu2021swin} encoder. Our implementation builds upon MoBY \cite{xie2021moby}, where the online encoder, $f_q$, consists of the backbone, a projection head \cite{chen2020simclr}, and an extra prediction head \cite{grill2020byol}; the target encoder, $f_k$, has the backbone and projection head, but does not include the prediction head. For training, we use the AdamW optimizer \cite{loshchilov2019adamw} with a base learning rate of $0.03$, weight decay of $10^{-4}$, and batch size of $512$. Unless otherwise stated, we use the following default hyperparameters. The momentum parameter starts at $m_\text{start}=0.99$ and increases to 1 following a cosine schedule. For synthetic negatives, we select the top $N=256$ hardest negatives. We use a temperature $\tau=0.2$ for the contrastive loss of \Cref{eq:loss_contrastive} and a queue size $K=4096$. For our experiments on ILSVRC-2012, we implement a cooldown period for the last 100 epochs where \textit{no} synthetic negatives are generated. For linear evaluation, we train a fully-connected layer on frozen features for 100 epochs. We refer the reader to \cite{xie2021moby} for further implementation details.

\subsection{Linear Evaluation on ImageNet}

\Cref{tab:lineval} shows top-1 accuracy of our method after pretraining for 300 epochs on ImageNet ILSVRC-2012. \NAME{} outperforms the MoBY baseline by 0.2\% on both architectures and maintains superiority over DINO \cite{caron2021dino} and MoCo-v3 \cite{chen2021mocov3}. However, we note that there is a considerable gap in performance when one compares with supervised training.

\begin{table}[!t]
\centering
\caption{Top-1 classification accuracy on ImageNet for self-supervised methods with DeiT-S and Swin-T architectures.}
\begin{tabular}{lccc}
\toprule
Method & Arch. & Params (M) & Top-1 (\%) \\
\midrule
\multirow{2}{*}{\textit{Supervised}} & DeiT-S & 22 & 79.8 \\
& Swin-T & 29 & 81.3 \\
\midrule
DINO \cite{caron2021dino} & DeiT-S & 22 & 72.5 \\
MoCo-v3 \cite{chen2021mocov3} & DeiT-S & 22 & 72.5 \\
\midrule
\multirow{2}{*}{MoBY \cite{xie2021moby}} & DeiT-S & 22 & 72.8 \\
& Swin-T & 29 & 75.0 \\
\midrule
\multirow{2}{*}{\NAME{} (ours)} & DeiT-S & 22 & \textbf{73.0} \\
& Swin-T & 29 & \textbf{75.2} \\
\bottomrule
\end{tabular}
\label{tab:lineval}
\end{table}

\paragraph{Visualizing attention.}

\Cref{fig:teaser} shows self-attention patterns of MoBY and \NAME{}. Though similar, ours tends to produce more focused attention maps with finer-grained patterns that highlight semantically meaningful regions. This suggests that synthetic negatives help develop more discriminative features, targeting relevant visual elements.

\subsection{Ablation Study}

Our ablation studies of \NAME{} are performed on ImageNet-100 classification, with pretraining for 100 epochs.

%%%%%%%%%%%%%%%%%%%%%%%%%%%%%%%%%%%%%%%%%% TABLE 2-3
\begin{table*}[!tb]
\centering
\begin{minipage}{0.48\textwidth}
  \centering
  \caption{Ablation study on applying tricks of MoCo-v3.}
  \begin{tabular}{cccc}
    \toprule
    Fixed Patch & Replace LN before & \multicolumn{2}{c}{Top-1 (\%)} \\
    \cmidrule(lr){3-4}
    Embedding & MLP with BN & \multicolumn{1}{c}{DeiT-S} & \multicolumn{1}{c}{Swin-T} \\
    \midrule
     &  & 66.7 & 67.5 \\
     \checkmark &  & 66.4 & 67.2 \\
     & \checkmark & 67.2 & 67.9 \\
    \bottomrule
  \end{tabular}
  \label{tab:ablation_mocov3}
\end{minipage}
\hfill
\begin{minipage}{0.48\textwidth}
  \centering
  \caption{Ablation study on the drop path rates.}
  \begin{tabular}{cccc}
    \toprule
    \emph{Online} & \emph{Target} & \multicolumn{2}{c}{Top-1 (\%)} \\
    \cmidrule(lr){3-4}
    dpr & dpr & \multicolumn{1}{c}{DeiT-S} & \multicolumn{1}{c}{Swin-T} \\
    \midrule
    %0.0 & 0.0 & 66.7 & 67.5 \\
    0.1  & 0.1 & 61.9 & 74.3 \\
    0.05 & 0.0 & 65.0 & 75.3 \\
    0.1  & 0.0 & 65.0 & 75.4 \\
    0.2  & 0.0 & 64.7 & 72.7 \\
    \bottomrule
  \end{tabular}
  \label{tab:ablation_odp}
\end{minipage}
\end{table*}

%%%%%%%%%%%%%%%%%%%%%%%%%%%%%%%%%%%%%%%%%% TABLE 4-6
\begin{table*}[!tb]
\centering
\begin{minipage}{0.32\textwidth}
    \centering
    \caption{Ablation study on queue size $K$.}
    \label{tab:ablation_queue}
    \begin{tabular}{ccc}
    \toprule
    \multirow{2}{*}{$K$} & \multicolumn{2}{c}{Top-1 (\%)} \\
    \cmidrule(lr){2-3}
     & DeiT-S & Swin-T \\
    \midrule
    1024 & 64.5 & 72.5 \\
    2048 & 64.5 & 72.5 \\
    4096 & 64.7 & 72.7 \\
    8192 & 63.6 & 72.3 \\
    16384 & 62.6 & 71.6 \\
    \bottomrule
    \end{tabular}
\end{minipage}
\hfill
\begin{minipage}{0.32\textwidth}
    \centering
    \caption{Ablation study on temperature $\tau$.}
    \label{tab:ablation_temp}
    \begin{tabular}{ccc}
    \toprule
    \multirow{2}{*}{$\tau$} & \multicolumn{2}{c}{Top-1 (\%)} \\
    \cmidrule(lr){2-3}
     & DeiT-S & Swin-T \\
    \midrule
    0.07 & 59.3 & 61.5 \\
    0.1 & 61.5 & 69.2 \\
    0.2 & 64.5 & 72.7 \\
    0.3 & 64.0 & 71.7 \\
    \bottomrule
    \end{tabular}
\end{minipage}
\hfill
\begin{minipage}{0.32\textwidth}
    \centering
    \caption{Ablation study on momentum $m_{\text{start}}$.}
    \label{tab:ablation_momentum}
    \begin{tabular}{ccc}
    \toprule
    \multirow{2}{*}{$m_{\text{start}}$} & \multicolumn{2}{c}{Top-1 (\%)} \\
    \cmidrule(lr){2-3}
     & DeiT-S & Swin-T \\
    \midrule
    0.99 & 64.5 & 72.7 \\
    0.993 & 65.2 & 72.2 \\
    0.996 & 63.8 & 72.4 \\
    0.999 & 60.3 & 68.6 \\
    \bottomrule
    \end{tabular}
\end{minipage}
\end{table*}

\paragraph{Synthetic negatives.} 

We observe architectural differences in how DeiT and Swin transformers respond to synthetic negatives, shown in \Cref{fig:ablation_hardness}. DeiT benefits from mining negatives at either low (256) or high (1024) hardness levels, while Swin performs well across all hardness levels. Additionally, DeiT achieves better results with moderately hard negatives at medium or high proportions whereas Swin performs consistently well with all proportions. This likely stems from Swin's inductive biases requiring less aggressive negative samples than DeiT's pure transformer architecture.

\paragraph{Applying MoCo-v3 tricks.} 

Our experiments reveal that synthetic negatives eliminate the need for additional stabilization techniques from MoCo-v3. As shown in \Cref{tab:ablation_mocov3}, fixing the patch embedding leads to worse results, suggesting that our synthetic negatives already provide sufficient regularization. This allows for a simpler implementation without compromising performance. Notably, replacing Layer Normalization (LN) with Batch Normalization (BN) before MLP blocks yields improvements.

\paragraph{Asymmetric drop path rates.} 

As shown in \Cref{tab:ablation_odp}, the asymmetric configuration of drop path rates (dpr) significantly impacts model performance. Unlike MoBY, which uses 0.2 for the online encoder, we find a smaller rate of 0.1 is optimal when combined with synthetic negatives. This is in agreement with our intuition that the synthetic negatives provide additional regularization and reduce the need for stablization tricks. Applying drop path only to the online encoder while keeping the target encoder stable yields the best balance.

\paragraph{Other hyperparameters.}

The default hyperparameters from MoBY work effectively with our synthetic negative approach. As shown in \Cref{tab:ablation_queue,tab:ablation_temp,tab:ablation_momentum}, performance remains stable across different queue sizes (best at 4096), temperatures (optimal at 0.2), and momentum values (best at 0.99). This demonstrates that synthetic negatives can be incorporated without extensive re-tuning of existing parameters. Overall, our synthetic negative generation technique integrates seamlessly with established contrastive learning frameworks, requiring minimal adaptation effort.

%%%%%%%%%%%%%%%%%%%%%%%%%%%%%%%%%%%%%%%%%%%%%%%%%%%%%%%%%%%%%%%%%%%%%%%%%%%%%%%
%%%%%%%%%%%%%%%%%%%%%%%%%%%%%%%%%%%%%%%%%%%%%%%%%%%%%%%%%%%%%%%%%%%%%%%%%%%%%%%
% 5 CONCLUSION
%%%%%%%%%%%%%%%%%%%%%%%%%%%%%%%%%%%%%%%%%%%%%%%%%%%%%%%%%%%%%%%%%%%%%%%%%%%%%%%
%%%%%%%%%%%%%%%%%%%%%%%%%%%%%%%%%%%%%%%%%%%%%%%%%%%%%%%%%%%%%%%%%%%%%%%%%%%%%%%

\section{Discussion} \label{sec:conclusion}

In this paper, we explored synthetic negatives in vision transformer pretraining. We found that synthetic negatives provide sufficient regularization for learning more robust representations while also reducing the need for stabilization tricks. Importantly, our approach requires minimal adjustments to current frameworks, working in a "\textit{plug-and-play}" manner with existing architectures. The experimental results demonstrate that \NAME{} further improves unsupervised learning with minimal computational overhead, showing consistent gains across different transformer architectures.

\paragraph{Limitations.} Our ablation studies were conducted on ImageNet-100, which may not fully capture the behavior on larger-scale datasets. Additionally, while we demonstrated improved performance on image classification, we did not evaluate on more complex downstream tasks.

\paragraph{Future work.} Synthetic hard negatives have proven effective for vision transformers. Exploring their integration into vision-language frameworks represents a promising direction, with potential to enhance cross-modal contrastive learning through more challenging negative examples.
%%%%%%%%%%%%%%%%%%%%%%%%%%%%%%%%%%%%%%%%%%%%%%%%%%%%%%%%%%%%%%%%%%%%%%%%%%%%%%%
%%%%%%%%%%%%%%%%%%%%%%%%%%%%%%%%%%%%%%%%%%%%%%%%%%%%%%%%%%%%%%%%%%%%%%%%%%%%%%%
% Acknowledgments
%%%%%%%%%%%%%%%%%%%%%%%%%%%%%%%%%%%%%%%%%%%%%%%%%%%%%%%%%%%%%%%%%%%%%%%%%%%%%%%
%%%%%%%%%%%%%%%%%%%%%%%%%%%%%%%%%%%%%%%%%%%%%%%%%%%%%%%%%%%%%%%%%%%%%%%%%%%%%%%

\section*{Acknowledgments}

We acknowledge the computational resources and support provided by the Imperial College Research Computing Service (\url{http://doi.org/10.14469/hpc/2232}), which enabled our experiments.

%\section*{Broader Impact}

%The improvements demonstrated by \NAME{} have potential implications beyond the specific task of image classification. Our approach contributes to more data-efficient deep learning. This is particularly valuable in domains where labeled data is scarce or expensive to obtain.

%%%%%%%%%%%%%%%%%%%%%%%%%%%%%%%%%%%%%%%%%%%%%%%%%%%%%%%%%%%%%%%%%%%%%%%%%%%%%%%
%%%%%%%%%%%%%%%%%%%%%%%%%%%%%%%%%%%%%%%%%%%%%%%%%%%%%%%%%%%%%%%%%%%%%%%%%%%%%%%
% References
%%%%%%%%%%%%%%%%%%%%%%%%%%%%%%%%%%%%%%%%%%%%%%%%%%%%%%%%%%%%%%%%%%%%%%%%%%%%%%%
%%%%%%%%%%%%%%%%%%%%%%%%%%%%%%%%%%%%%%%%%%%%%%%%%%%%%%%%%%%%%%%%%%%%%%%%%%%%%%%

{
    \small
    \bibliographystyle{ieeenat_fullname}
    \bibliography{example_bibliography}
}

%%%%%%%%%%%%%%%%%%%%%%%%%%%%%%%%%%%%%%%%%%%%%%%%%%%%%%%%%%%%%%%%%%%%%%%%%%%%%%%
%%%%%%%%%%%%%%%%%%%%%%%%%%%%%%%%%%%%%%%%%%%%%%%%%%%%%%%%%%%%%%%%%%%%%%%%%%%%%%%
% END
%%%%%%%%%%%%%%%%%%%%%%%%%%%%%%%%%%%%%%%%%%%%%%%%%%%%%%%%%%%%%%%%%%%%%%%%%%%%%%%
%%%%%%%%%%%%%%%%%%%%%%%%%%%%%%%%%%%%%%%%%%%%%%%%%%%%%%%%%%%%%%%%%%%%%%%%%%%%%%%

\end{document}

%% file: preamble.tex
\usepackage{wrapfig}
\usepackage{url}
\usepackage{booktabs}
\usepackage{amsfonts}
\usepackage{nicefrac}
\usepackage{microtype}
\usepackage{lipsum}
\usepackage{fancyhdr} 
\usepackage{graphicx}
\usepackage{appendix}
\usepackage{algorithmic}
\usepackage{color,soul}
\sethlcolor{yellow}
\usepackage{multirow}
\usepackage{amsmath}
\usepackage{listings}
\usepackage{verbatim}
\usepackage{float}
\usepackage{subcaption}
\usepackage{comment}
\usepackage{dsfont}
\usepackage{algorithm}
\usepackage{booktabs}
\usepackage{caption} 
\usepackage{xcolor}
\usepackage{colortbl} 
\usepackage{threeparttable}
\definecolor{darkgreen}{RGB}{0,190,0}
\usepackage{amssymb}
\usepackage{pifont}
\usepackage{tikz}
\usetikzlibrary{decorations.pathreplacing, positioning}
\definecolor{mygreen}{RGB}{130,179,102}
\definecolor{myblue}{RGB}{108,142,191}
\definecolor{cvprorange}{RGB}{217,95,2} % A strong CVPR-style orange
\usepackage{tipa}

\usepackage{etoolbox}
\makeatletter
\AfterEndEnvironment{algorithm}{\let\@algcomment\relax}
\AtEndEnvironment{algorithm}{\kern2pt\hrule\relax\vskip3pt\@algcomment}
\let\@algcomment\relax
\newcommand\algcomment[1]{\def\@algcomment{\footnotesize#1}}
\renewcommand\fs@ruled{\def\@fs@cfont{\bfseries}\let\@fs@capt\floatc@ruled
  \def\@fs@pre{\hrule height.8pt depth0pt \kern2pt}%
  \def\@fs@post{}%
  \def\@fs@mid{\kern2pt\hrule\kern2pt}%
  \let\@fs@iftopcapt\iftrue}
\makeatother